\documentclass[11pt]{article}
\usepackage{eacl2017}
\usepackage{times}
\usepackage{latexsym}
\usepackage{amsmath}
\usepackage{amssymb}
\usepackage{graphicx}
\usepackage{subfigure}
\usepackage{epsfig}
\usepackage{epstopdf}
\usepackage{url}
\usepackage{latexsym}
\DeclareMathOperator*{\argmax}{arg\,max}

\eaclfinalcopy % Uncomment this line for the final submission
\title{F-Score Driven Max Margin Neural Network for Named Entity Recognition in Chinese Social Media}

\author{Hangfeng He \and Xu Sun\\
	MOE Key Laboratory of Computational Linguistics, Peking University\\
	School of Electronics Engineering and Computer Science, Peking University\\
	\{hangfenghe, xusun\} @pku.edu.cn\\
}

\date{}

\begin{document}

	\maketitle

	\begin{abstract}
		We focus on named entity recognition (NER) for Chinese social media. With massive unlabeled text and quite limited labelled corpus, we propose a semi-supervised learning model based on B-LSTM neural network. To take advantage of traditional methods in NER such as CRF, we combine transition probability with deep learning in our model. To bridge the gap between label accuracy and F-score of NER, we construct a model which can be directly trained on F-score. When considering the instability of F-score driven method and meaningful information provided by label accuracy, we propose an integrated method to train on both F-score and label accuracy. Our integrated model yields substantial improvement over previous state-of-the-art result.
		
	\end{abstract}

	\section{Introduction}
	With the development of Internet, social media plays an important role in information exchange. The natural language processing tasks on social media are more challenging which draw attention of many researchers ~\cite{li-liu:2015:ACL-IJCNLP,habib-vankeulen:2015:ACL-IJCNLP-2015-System-Demonstrations,radford-carreras-henderson:2015:EMNLP,cherry-guo:2015:NAACL-HLT}. As the foundation of many downstream applications \cite{weissenborn-EtAl:2015:ACL-IJCNLP,delgado-EtAl:2014:Coling,hajishirzi-EtAl:2013:EMNLP} such as information extraction, named entity recognition (NER) deserves more research in prevailing and challenging social media text. NER is a task to identify names in texts and to assign names with particular types \cite{SunIJCAI09,Sun_NIPS2014,SunLWL14,HeAAAI17}. It is the informality of social media that discourages accuracy of NER systems. While efforts in English have narrowed the gap between social media and formal domains \cite{cherry-guo:2015:NAACL-HLT}, the task in Chinese remains challenging. It is caused by Chinese logographic characters which lack many clues to indicate whether a word is a name, such as capitalization. The scant labelled Chinese social media corpus makes the task more challenging \cite{Neelakantan2015Learning,Maria2014Enhancing,liu-EtAl:2015:NAACL-HLT4}.
	
	To address the problem, one approach is to use the lexical embeddings learnt from massive unlabeled text. To take better advantage of unlabeled text, Peng and Dredze~\shortcite{peng-dredze:2015:EMNLP} evaluates three types of embeddings for Chinese text, and shows the effectiveness of positional character embeddings with experiments. Considering the value of word segmentation in Chinese NER, another approach is to construct an integrated model to jointly train learned representations for both predicting word segmentations and NER~\cite{peng-dredze:2016:P16-2}.
	
	However, the two above approaches are implemented within CRF model. We construct a semi-supervised model based on B-LSTM neural network to learn from the limited labelled corpus by using lexical information provided by massive unlabeled text. To shrink the gap between label accuracy and F-Score, we propose a method to directly train on F-Score rather than label accuracy in our model. In addition, we propose an integrated method to train on both F-Score and label accuracy. Specifically, we make contributions as follows:
	\begin{itemize}
		\item We propose a method to directly train on F-Score rather than label accuracy. In addition, we propose an integrated method to train on both F-Score and label accuracy.
		
		\item We combine transition probability into our B-LSTM based max margin neural network to form structured output in neural network.
		
		\item We evaluate two methods to use lexical embeddings from unlabeled text in neural network.
	\end{itemize}

	\section{Model}
	
	We construct a semi-supervised model which is based on B-LSTM neural network and combine transition probability to form structured output. We propose a method to train directly on F-Score in our model. In addition, we propose an integrated method to train on both F-Score and label accuracy.
	
	\subsection{Transition Probability}
	\label{sect:transprob}
	B-LSTM neural network can learn from past input features and LSTM layer makes it more efficient \cite{hammerton2003named,hochreiter1997long,chen-EtAl:2015:EMNLP2,Graves2006Connectionist}. However, B-LSTM cannot learn sentence level label information. Huang et al.~\shortcite{huang2015bidirectional} combine CRF to use sentence level label information. We combine transition probability into our model to gain sentence level label information.
	To combine transition probability into B-LSTM neural network, we construct a Max Margin Neural Network (MMNN)~\cite{pei-ge-chang:2014:P14-1} based on B-LSTM. The prediction of label in position $t$ is given as:	\begin{equation}
		\label{equ:output}
		y_t = softmax(W_{hy} * h_t + b_y)
	\end{equation}
	where $W_{hy}$ are the transformation parameters, $h_t$ the hidden vector and $b_y$ the bias parameter. For a input sentence $c_{[1:n]}$ with a label sequence $l_{[1:n]}$, a sentence-level score is then given as:
	\begin{equation*}
		\label{equ:score}
		s(c_{[1:n]},l_{[1:n]},\theta) = \sum^n_{t=1} (A_{l_{t-1}l_t} + f_{\Lambda}(l_t |c_{[1:n]}))
	\end{equation*}
	where$f_{\Lambda}(l_t |c_{[1:n]})$ indicates the probability of label $l_t$ at position $t$ by the network with parameters $\Lambda$, $A$ indicates the matrix of transition probability. In our model, $f_{\Lambda}(l_t |c_{[1:n]})$ is computed as:
	\begin{equation}
		f_{\Lambda}(l_t |c_{[1:n]}) = -log(y_t[l_t])
	\end{equation}
	
	We define a structured margin loss $\Delta(l, \overline{l})$ as Pei et al.~\shortcite{pei-ge-chang:2014:P14-1}:
	\begin{equation}\Delta(l, \overline{l}) = \sum_{j=1}^n\kappa {\bf{1}} \{l_j \neq \overline{l}_j\}
	\end{equation}
	where $n$ is the length of setence $x$, $\kappa$ is a discount parameter, $l$ a given correct label sequence and $\overline{l}$ a predicted label sequence. For a given training instance $(x_i,y_i)$, our predicted label sequence is the label sequence with highest score:
	%$$\argmax_{i \in I} \varphi_i $$
	$$l^*_i = \argmax_{\overline{l}_i \in {Y(x_i)}}s(x_i, \overline{l}_i,\theta) $$
	The label sequence with the highest score can be obtained by carrying out viterbi algorithm.
	The regularized objective function is as follows:
	\begin{equation}
		\label{equ:costfunc}
		{J(\theta) = \frac{1}{m} \sum_{i=1}^m q_i(\theta) + \frac{\lambda}{2}||\theta||^2}
	\end{equation}
	\begin{equation*}
		q_i(\theta) = \max \limits_{\overline{l}_i \in Y(x_i)}(s(x_i,\overline{l}_i,\theta) + \Delta({l_i,\overline{l}_i})) - s(x_i, l_i, \theta)
	\end{equation*}
	By minimizing the object, we can increase the score of correct label sequence $l$ and decrease the score of incorrect label sequence $\overline{l}$.
	\begin{figure*}
		
		\centering
		\subfigure[F-Score of the models.]{
			\includegraphics[scale=0.33]{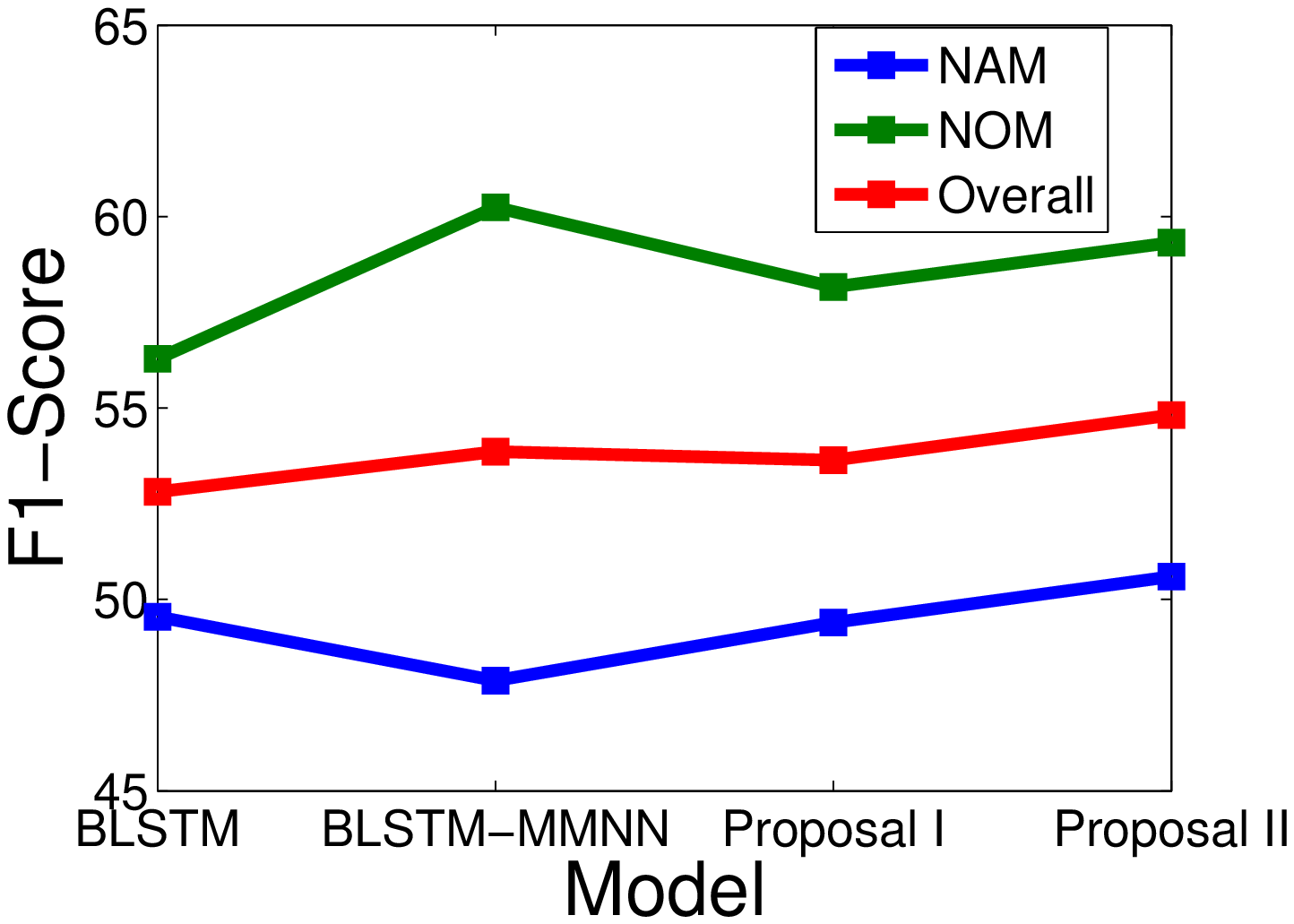}
			
			\label{fig:fscore}}
		\hspace{0.1in}
		\subfigure[Running time of the models.]{
			\centering
			\includegraphics[scale=0.33]{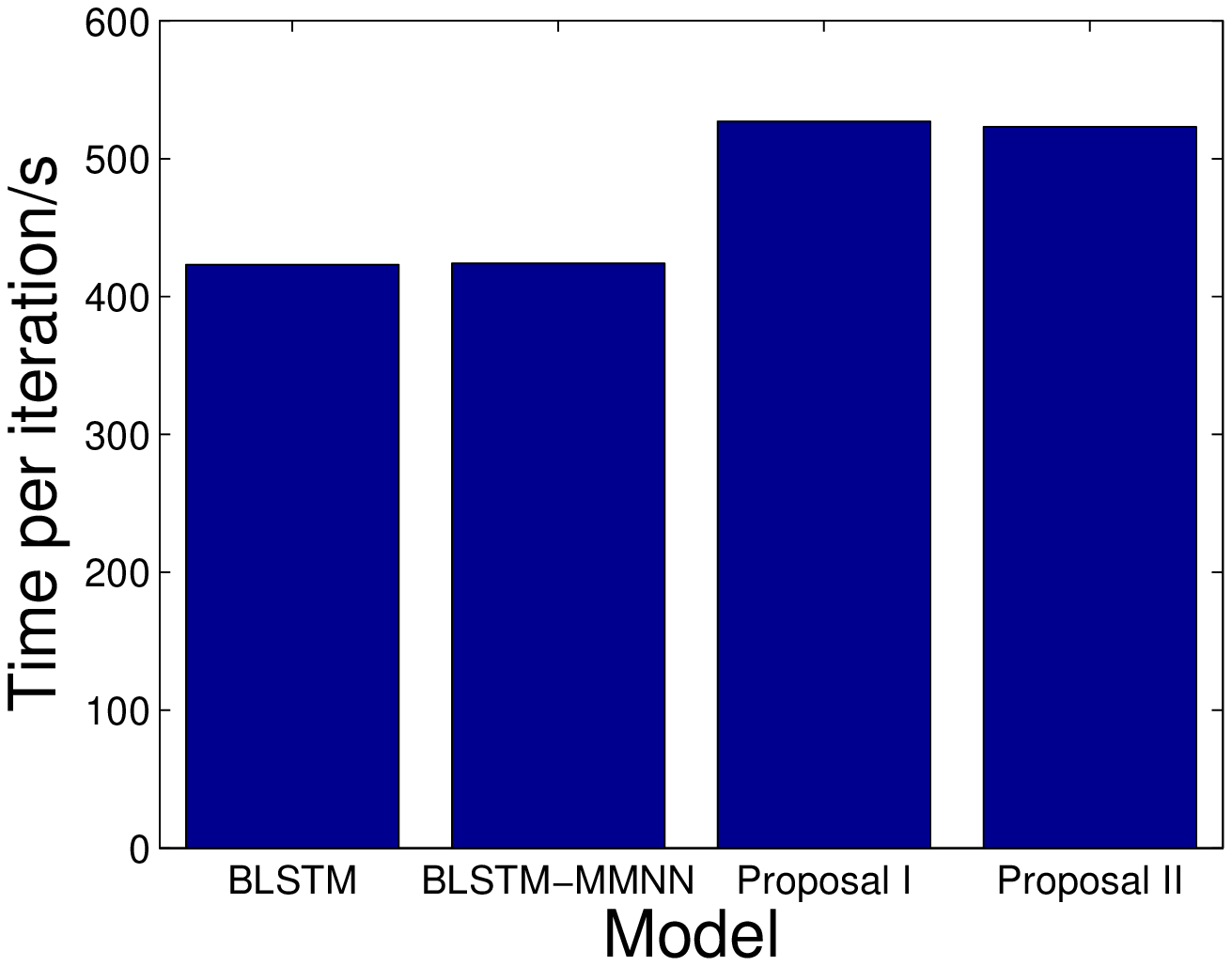}
			
			\label{fig:time}}
		\subfigure[Overall F1-Score with different values of beta.]{
			\centering
			\includegraphics[scale=0.33]{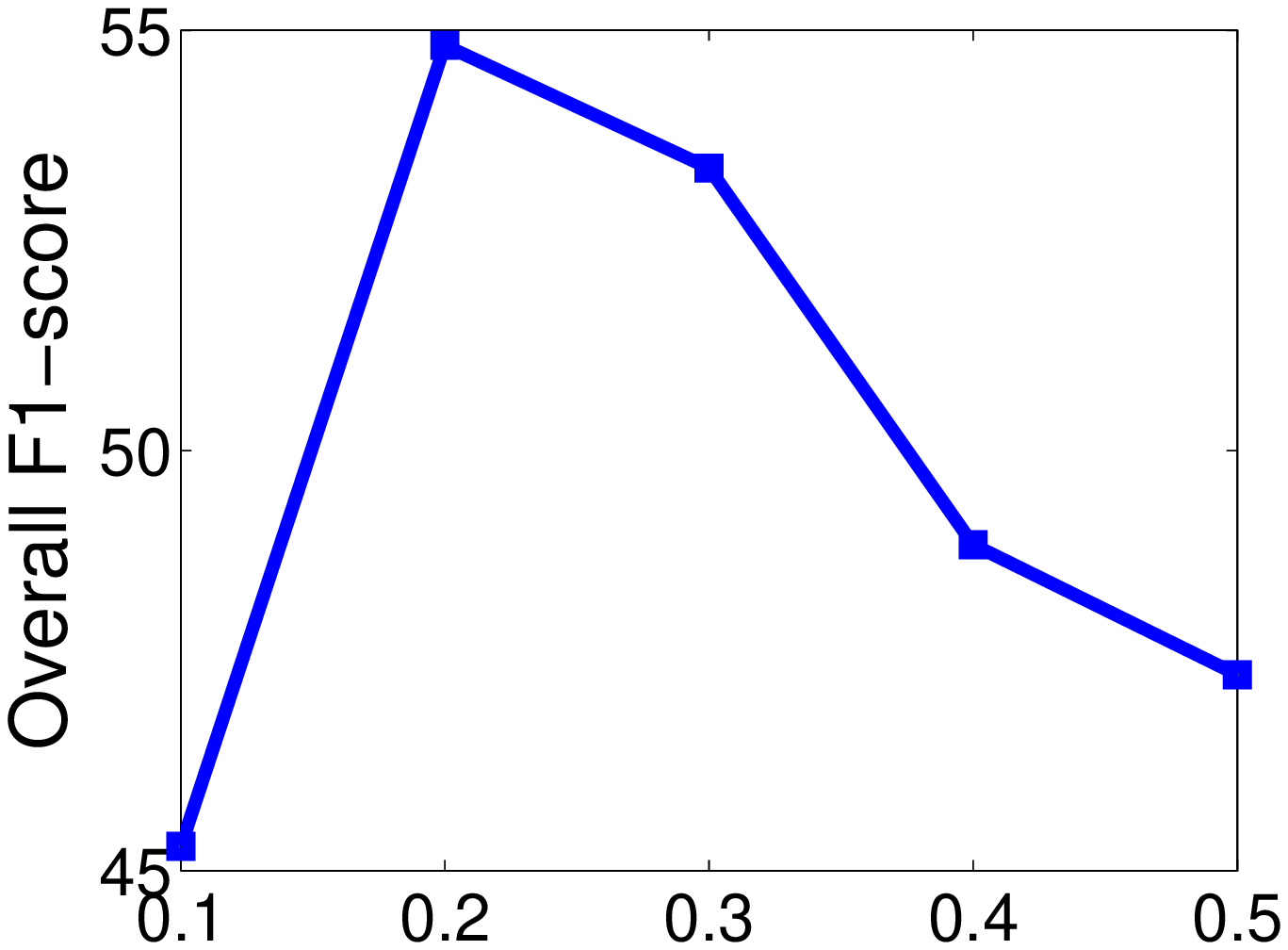}
			
			\label{fig:beta}}
		%\caption{Comparing different models.}
		%\label{fig:total}

	\end{figure*}
	
	\subsection{F-Score Driven Training Method}
	\label{f1}
	Max Margin training method use structured margin loss $\Delta(l, \overline{l})$ to describe the difference between the corrected label sequence $l$ and predicted label sequence $\overline{l}$. In fact, the structured margin loss $\Delta(l,\overline{l})$ reflect the loss in label accuracy. Considering the gap between label accuracy and F-Score in NER, we introduce a new training method to train directly on F-Score. To introduce F-Score driven training method, we need to take a look at the subgradient of equation (\ref{equ:costfunc}):
	\begin{equation*}
		\frac{\partial J}{\partial \theta} = \frac{1}{m}\sum_{i=1}^m(\frac{\partial s(x, \overline{l}_{max}, \theta)}{\partial\theta} - \frac{\partial s(x, l, \theta)}{\partial\theta}) + \lambda\theta
	\end{equation*}
	In the subgradient, we can know that structured margin loss $\Delta(l, \overline{l})$ contributes nothing to the subgradient of the regularized objective function $J(\theta)$. The margin loss $\Delta(l,\overline{l})$ serves as a trigger function to conduct the training process of B-LSTM based MMNN. We can introduce a new trigger function to guide the training process of neural network.
	
	\noindent{\bf F-Score Trigger Function} The main criterion of NER task is F-score. However, high label accuracy does not mean high F-score. For instance, if every named entity's last character is labeledas O, the label accuracy can be quite high, but the precision, recall and F-score are $0$. We use the F-Score between corrected label sequence and predicted label sequence as trigger function, which can conduct the training process to optimize the F-Score of training examples. Our new structured margin loss can be described as:
	\begin{equation}
		\widetilde{\Delta}(l,\overline{l}) = \kappa * FScore
	\end{equation}
	where $FScore$ is the F-Score between corrected label sequence and predicted label sequence.
	
	\noindent{\bf F-Score and Label Accuracy Trigger Function} The F-Score can be quite unstable in some situation. For instance, if there is no named entity in a sentence, F-Score will be always $0$ regardless of the predicted label sequence. To take advantage of meaningful information provided by label accuracy, we introduce an integrated trigger function as follows:
	\begin{equation}
		\hat{\Delta}(l,\overline{l}) = \widetilde{\Delta}(l,\overline{l}) + \beta * \Delta(l, \overline{l})
	\end{equation}
	where $\beta$ is a factor to adjust the weight of label accuracy and F-Score.
	
	Because F-Score depends on the whole label sequence, we use beam search to find $k$ label sequences with top sentece-level score $s(x,\overline{l},\theta)$ and then use trigger function to rerank the $k$ label sequences and select the best.

	\subsection{Word Segmentation Representation}
	\label{sect:wordseg}
	Word segmentation takes an important part in Chinese text processing. Both Peng and Dredze~\shortcite{peng-dredze:2015:EMNLP} and Peng and Dredze~\shortcite{peng-dredze:2016:P16-2} show the value of word segmentation to Chinese NER in social media. We present two methods to use word segmentation information in neural network model.
	
	\noindent{\bf Character and Position Embeddings} To incorporate word segmentation information, we attach every character with its positional tag. This method is to distinguish the same character at different position in the word. We need to word segment the text and learn positional character embeddings from the segmented text.
	
	\noindent{\bf Character Embeddings and Word Segmentation Features}
	We can treat word segmentation as discrete features in neural network model. The discrete features can be easily incorporated into neural network model~\cite{collobert2011natural}. We use word embeddings from a LSTM pretrained on MSRA 2006 corpus to initialize the word segmentation features.

	\section{Experiments and Analysis}
	\label{sec:exper}

	\subsection{Datasets}
	\begin{table}[h]
		\centering
		\begin{tabular}{|c|c|c|}
			\hline
			& \textbf{Named}&\textbf{Nominal}  \\ \hline
			Train set & 957& 898    \\ \hline
			Development set & 153& 226 \\ \hline
			Test set & 209& 196\\ \hline
			Unlabeled Text&         \multicolumn{2}{c|}{112,971,734  Weibo messages}               \\ \hline
		\end{tabular}
		\caption{Details of Weibo NER corpus.}
		\label{table:data}
	\end{table}
	We use a modified labelled corpus\footnote{We fix some labeling errors of the data.} as Peng and Dredze~\shortcite{peng-dredze:2016:P16-2} for NER in Chinese social media. Details of  the data are listed in Table \ref{table:data}. We also use the same unlabelled text as Peng and Dredze~\shortcite{peng-dredze:2016:P16-2} from Sina Weibo service in China and the text is word segmented by a Chinese word segmentation system Jieba\footnote{https://github.com/fxsjy/jieba.} as Peng and Dredze~\shortcite{peng-dredze:2016:P16-2} so that our results are more comparable to theirs.
	\begin{table*}
		\centering
		\begin{tabular}{|c|c|c|c|c|c|c|}
			\hline
			\multicolumn{1}{|c|}{\textbf{Methods}} & \multicolumn{3}{c}{\textbf{Named Entity}}                                                                                                     & \multicolumn{3}{|c|}{\textbf{Nominal Mention}}                                                                                                      \\ \hline
			& \multicolumn{1}{c|}{\textbf{Precision}} & \multicolumn{1}{c|}{\textbf{Recall}} & \multicolumn{1}{c|}{\textbf{F1}} & \multicolumn{1}{c}{\textbf{Precision}} & \multicolumn{1}{|c}{\textbf{Recall}} & \multicolumn{1}{|c|}{\textbf{F1}} \\ \hline
			Character+Segmentation & 48.52 & 39.23 & 43.39 & 58.75 & \textbf{47.96} & 52.91  \\ \hline
			Character+Position     & \textbf{65.87} & \textbf{39.71} & \textbf{49.55} & \textbf{68.12} & \textbf{47.96} & \textbf{56.29}  \\ \hline
		\end{tabular}
		\caption{Two methods to incorporate word segmentation information.}
		\label{table:seg}
	\end{table*}
	\begin{table*}
		\centering
		\scalebox{0.9}{
		\begin{tabular}{|c|c|c|c|c|c|c|c|c|}
			\hline
			\multicolumn{1}{|c|}{\textbf{Models}} & \multicolumn{3}{c}{\textbf{Named Entity}}                                                                                                     & \multicolumn{3}{|c|}{\textbf{Nominal Mention}}
			&\multicolumn{2}{|c|}{}                                           \\ \hline
			& \multicolumn{1}{c|}{\textbf{Precision}} & \multicolumn{1}{c|}{\textbf{Recall}} & \multicolumn{1}{c|}{\textbf{F1}} & \multicolumn{1}{c}{\textbf{Precision}} & \multicolumn{1}{|c}{\textbf{Recall}} & \multicolumn{1}{|c|}{\textbf{F1}} &
			\multicolumn{1}{c}{\textbf{Overall}} & \multicolumn{1}{|c|}{\textbf{OOV}}
			\\ \hline
			\cite{peng-dredze:2015:EMNLP}     & 57.98 & 35.57 & 44.09 & 63.84 & 29.45 & 40.38 & 42.70 & -\\ \hline
			\cite{peng-dredze:2016:P16-2}     & 63.33 & 39.18 & 48.41 & 58.59 & 37.42 & 45.67 & 47.38 & - \\ \hline
			B-LSTM     & 65.87 & 39.71 & 49.55 & 68.12 & 47.96 & 56.29 & 52.81 & 13.97 \\ \hline
			B-LSTM + MMNN & 65.29 & 37.80 & 47.88 & \textbf{73.53} & 51.02 & \textbf{60.24}  & 53.86 & 17.90\\ \hline
			F-Score Driven \uppercase\expandafter{\romannumeral1} (proposal)   & 66.67 & 39.23 & 49.40 & 69.50 & 50.00 & 58.16 & 53.64 & 17.03 \\ \hline
			F-Score Driven \uppercase\expandafter{\romannumeral2} (proposal)   & \textbf{66.93} & \textbf{40.67} & \textbf{50.60} & 66.46 & \textbf{53.57} & 59.32 & \textbf{54.82} & \textbf{20.96}\\ \hline
		\end{tabular}}
		\caption{NER results for named and nominal mentions on test data.}
		\label{table:model}
	\end{table*}
	
	\subsection{Parameter Estimation}
	We pre-trained embeddings using word2vec ~\cite{mikolov2013distributed} with the skip-gram training model, without negative sampling and other default parameter settings. Like Mao et al.~\shortcite{mao2008chinese}, we use bigram features as follow:
	$$C_nC_{n+1}(n=-2,-1,0,1) \quad and \quad C_{-1}C_1$$
	We use window approach~\cite{collobert2011natural} to extract higher level Features from word feature vectors. We treat bigram features as discrete features ~\cite{collobert2011natural} for our neural network. Our models are trained using stochastic gradient descent with an L2 regularizer.
	\\
	As for parameters in our models, window size for word embedding is $5$, word embedding dimension, feature embedding dimension and hidden vector dimension are all $100$, discount $\kappa$ in margin loss is $0.2$, and the hyper parameter for the $L2$ is $0.000001$. As for learning rate, initial learning rate is $0.1$ with a decay rate $0.95$. For integrated model, $\beta$ is $0.2$. We train $20$ epochs and choose the best prediction for test.
	
	\subsection{Results and Analysis}
	%\begin{figure}
	
	%	\centering
	
	%   \includegraphics[scale=0.4]{beta.eps}
	
	%	\caption{Overall F1-Score with different values of beta.}
	%	\label{fig:beta}

	%\end{figure}
	We evaluate two methods to incorporate word segmentation information. The results of two methods are shown as Table \ref{table:seg}. We can see that positional character embeddings perform better in neural network. This is probably because positional character embeddings method can learn word segmentation information from unlabeled text while word segmentation can only use training corpus.

	We adopt positional character embeddings in our next four models. Our first model is a B-LSTM neural network (baseline). To take advantage of traditional model \cite{chieu2002named,Mccallum2001Maximum} such as CRF, we combine transition probability in our B-LSTM based MMNN. We design a F-Score driven training method in our third model F-Score Driven Model \uppercase\expandafter{\romannumeral1} . We propose an integrated training method in our fourth model F-Score Driven Model \uppercase\expandafter{\romannumeral2} .The results of models are depicted as Figure \ref{fig:fscore}. From the figure, we can know our models perfrom better with little loss in time.
	
	Table \ref{table:model} shows results for NER on test sets. In the Table \ref{table:model}, we also show  micro F1-score (Overall) and out-of-vocabulary entities (OOV) recall. Peng and Dredze~\shortcite{peng-dredze:2016:P16-2} is the state-of-the-art NER system in Chinese Social media. By comparing the results of B-LSTM model and B-LSTM + MTNN model, we can know transition probability is significant for NER. Compared with B-LSTM + MMNN model, F-Score Driven Model
	\uppercase\expandafter{\romannumeral1} improves the result of named entity with a loss in nominal mention.
	%As for the loss in nominal mention, it may be caused by the sentences without a named entity or nominal mention. The detailed analysis can be found in section \ref{f1}.
	The integrated training model (F-Score Driven Model
	\uppercase\expandafter{\romannumeral2}) benefits from both label accuracy and F-Score, which achieves a new state-of-the-art NER system in Chinese social media. Our integrated model has better performance on named entity and nominal mention.
	
	To better understand the impact of the factor $\beta$, we show the results of our integrated model with different values of $\beta$ in Figure \ref{fig:beta}. From Figure \ref{fig:beta}, we can know that $\beta$ is an important factor for us to balance F-score and accuracy. Our integrated model may help alleviate the influence of noise in NER in Chinese social media.
	
	\section{Conclusions and Future Work}
	\label{sec:discuss}
	The results of our experiments also suggest directions for future work. We can observe all models in Table \ref{table:model} achieve a much lower recall than precision \cite{pink-nothman-curran:2014:EMNLP2014}. So we need to design some methods to solve the problem.
	
	\section*{Acknowledgements}
	Thanks to Shuming Ma for the help on improving the writing. This work was supported in part by National Natural Science Foundation of China (No. 61673028), and National High Technology Research and Development Program of China (863 Program, No. 2015AA015404). Xu Sun is the corresponding author of this paper. The first author focuses on the design of the method and the experimental results. The corresponding author focuses on the design of the method.
	
	\bibliography{eacl2017}
	\bibliographystyle{eacl2017}

\end{document}